\documentclass[journal]{IEEEtran}

\ifCLASSINFOpdf

\else

\fi

\usepackage{multirow}

\usepackage{amsmath}

\usepackage{graphicx}

\usepackage{color}

\begin{document}

\title{Cross-database non-frontal facial expression recognition based on transductive deep transfer learning}


\author{Keyu Yan,
        Wenming~Zheng*,~\IEEEmembership{Member,~IEEE,}
        Tong Zhang,
        Yuan~Zong
        and Zhen Cui
\thanks{ Keyu Yan and Tong Zhang are with the Key Laboratory of Child Development and Learning Science of Ministry of Education, and the Department
of Information Science and  Engineering, Southeast University, China. \protect\\
E-mail: yankeyu@seu.edu.cn.}
\thanks{Wenming Zheng and Yuan Zong and Jingwei Yan are with the Key Laboratory of Child Development and Learning Science of Ministry of Education, Research Center for Learning Science,
Southeast University, Nanjing, Jiangsu 210096, China.\protect\\
E-mail: wenming\_zheng@seu.edu.cn.}
\thanks{Zhen Cui is with School of Computer Science and Engineering, Nanjing University of Science and Technology, Nanjing, Jiangsu, 210096, China.\protect\\ 
Asterisk indicates corresponding author.}
}

\markboth{}%
{Shell \MakeLowercase{\textit{et al.}}: }

\maketitle

\begin{abstract}
 Cross-database non-frontal expression recognition is a very meaningful but rather difficult subject in the fields of computer vision and affect computing. In this paper, we proposed a novel transductive deep transfer learning architecture based on widely used  VGGface16-Net for this problem. In this framework, the VGGface16-Net is used to jointly learn an common optimal nonlinear discriminative features from the non-frontal facial expression samples between the source and target databases and then we design a novel transductive transfer layer to deal with the cross-database non-frontal facial expression classification task. In order to validate the performance of the proposed transductive deep transfer learning networks, we present extensive cross-database experiments on two famous available facial expression databases, namely the BU-3DEF and the Multi-PIE database. The final experimental results show that our transductive deep transfer network outperforms the state-of-the-art cross-database facial expression recognition methods.
\end{abstract}


%
\IEEEpeerreviewmaketitle

\section{Introduction}
\label{sec:intro}
Recently, artificial intelligence (AI) technology has made explosive progress in many practical applications such as driverless car, human-computer interaction, school education, intelligent transportation, et al. However, the success of AI technology at present is only based on a large number of labeled data. In fact, many real scenarios can not be expressed by data including creative driven learning, knowledge system and learning to learn, et al. These sophisticated problems require machine to understand and comprehend human modes and emotions. Therefore, how to make machine interpret human emotion will become the next much talked-about topic in the AI and machine learning community. Human express emotions in a variety of ways including language, facial expression, gesture, and word, in which the facial expression is the most important channel to convey emotion information between different people. Therefore, many AI researchers and computer technician have paid great attention to the facial expression recognition (FER) problem to help machine perception human emotion and have gained some harvest from it~\cite{ekman1978facial},~\cite{zhi2011graph},~\cite{dhall2017individual},~\cite{valstar2012meta},~\cite{nguyen2017deep},
~\cite{barsoum2016training},~\cite{Zong2018Learning},~\cite{zheng2006facial}. In order to study facial expression more conveniently and systematically, Ekman~\cite{ekman1978facial} identified six basis expressions across all cultures and defined the standard for facial expression research named the Facial Action Coding System (FACS). Moreover, Zhi et al.~\cite{zhi2011graph} proposed a novel non-negative matrix factorization method based on graph-preserving sparse (GSNMF)for facial expression recognition problem. The GSNMF algorithm acquires better representation by transforming the high-dimensional facial expression images into a locality preserving subspace with sparse representation and achieves higher recognition results than NMF. In ~\cite{nguyen2017deep}, Nguyen et al. proposed a multimodal approach to recognize dynamic facial expression by combining a 3-dimensional convolutional neural networks (C3Ds) which extract the spatio-temporal features and a deep-belief networks (DBN) which can represent audio and video streams for dynamic expression databases.

In practical FER applications, the same expression often can be acquired at different viewpoints, thus generating multi-view heterogeneous samples whose statistical information have a great difference. However, most of the traditional FER methods are based on the frontal facial expression samples and the non-frontal facial expression data are only a small part adopted. But despite all that, some novel methods are still employed to tackling the non-frontal FER problem and have made great progress in recent years~\cite{moore2011local},~\cite{zheng2014emotion},~\cite{tang2010non},~\cite{zhang2016deep},~\cite{zheng2009novel},~\cite{moore2010multi},~\cite{kumano2009pose}. In~\cite{tang2010non}, Tang et al.~\cite{tang2010non} built the ergodic hidden Markov model (EHMM) to obtain supervector representation of non-frontal facial expression images and achieve promising results. In~\cite{kumano2009pose} Kumano et al. propose a method which uses the variable-intensity template model to describe pose-invariant facial expressions from monocular video sequences. This method can estimate facial poses and expressions simultaneously by using a particle filter. In order to learn suitable optimal features for classifying the facial expressions from different facial view-points, Zhang et al.~\cite{zhang2016deep} proposed a feature-based deep neural network learning method which uses multiple network layers to describe the relationship between non-frontal facial features and their corresponding high-level semantic information.

Although the FER technology has achieved great success, however, the most of FER methods usually developed to based on the assumption of uniform probability distribution between the training and testing samples. In fact, in many practical scenarios, the hypothesis of uniform probability distribution is not satisfied because the training and testing data may come from two different databases which are acquired under the different environments or equipments. This leads to a challenging problem, namely, the cross-database non-frontal FER problem. To cope with this challenge, many effective approaches such as subspace-based methods and based on deep learning models had been proposed in recent years~\cite{zheng2015cross},~\cite{chu2017selective},~\cite{zong2018domain}~\cite{zheng2016cross},~\cite{wei2016deep},~\cite{zavarez2017cross},
~\cite{zhu2014weakly},~\cite{duan2012domain}. To leverage the distribution discrepancy between training and testing facial expression images, in our preliminary work in~\cite{zheng2016cross}, zheng et al. proposed a novel transductive transfer subspace learning method to jointly learn a discriminative subspace and to predict the label values of the unlabelled facial expression images by using all labelled training samples from source domain and an unlabelled auxiliary testing samples set from target domain. Duan et al. ~\cite{duan2012domain} propose a new cross-domain kernel learning framework named Domain Transfer Multiple Kernel Learning (DTMKL) to deal with the wide divergences between feature distributions of different databases. In~\cite{wei2016deep}, Wei et al. proposed a deep nonlinear feature coding framework for unsupervised cross-domain FER problem, which introduce domain divergence minimization by Maximum Mean Discrepancy (MMD) and kernelization coding to build on a marginalized stacked denoising auto-encoder for extracting very efficient deep features. Zavarez et al.~\cite{zavarez2017cross} utilize the fine-tune trick in deep convolutional network for cross-database video-based FER problem in several well-established facial expression databases. However, these methods of cross-database FER are typically based on the frontal and near-frontal facial samples or the samples of each domain with single view-point in their experiments. In many real application scenarios, FER not only faces with cross-database facial expression samples, but also handles a large number of non-frontal facial expression data, these facial expression images of different view-points also lead to the different distribution in the same database, making it more difficult to recognize the facial expression categories. Furthermore, when the non-frontal facial expression data are adopted in commonly cross-database FER task, it leads to a cross-database non-frontal FER problem which is a largely unexplored research field. This a very difficult subject, because researchers not only need to deal with the difference of distribution between databases, but also consider the distribution discrepancy under the same expression intra-databases, which leads to bigger difficult and challenge in learning the more discriminative facial expression features.

In this paper, we will address such a difficult and challenge problem, that is, cross-database non-frontal FER problem. For this purpose, we further expanding our preliminary work in~\cite{zheng2016cross} from linear to nonlinear method by dint of deep neural network model to propose a transductive deep transfer learning framework which a novel transfer network layer is introduced in this framework. Considering the eminent performance of VGGface16-Net in human facial feature representation, we first utilize the VGGface16-Net to learn the excellent representation of multi-view facial expression feature from the raw non-frontal facial expression images. Behind the VGGface16-Net framework, we design a novel transfer layer architecture for cross-database non-frontal facial expression classification task. In this task, the loss function and the network parameters are jointly optimized, and obtained the prediction label values of the target database samples. In summary, the main contributions of this paper are summarized as follows:
\begin{enumerate}
\item
Different from the traditional hand-craft features trick in facial expression recognition method, we utilize VGGface16-Net end to end to learn a common feature representation between source and target database. In the optimization phase of the TDTL network, the source and target databases are mapped to a common nonlinear feature representation space, and the discrepancy of distribution from non-frontal view-points features in the same database and the distribution different of between databases are eliminated as much as possible.
\item
In this paper, we designed a novel transductive transfer layer based on deep learning architecture to adaptively deal with cross-database non-frontal FER problem. We random initialize the labels of target database, meanwhile these label values were learn with labelled source database and the unlabelled target database in this network framework, and final to obtain the predicted label values of target database.
\item
Unlike the traditional subspace learning methods, in TDTL model the training and testing samples are divided into different batches to be jointly optimized in the network framework so as to better predict label values of the target samples.
\end{enumerate}
The remainder of the paper is structured as follows: Section 2 presents the transductive deep transfer learning method and shows how it run for cross-database non-frontal FER problem. The details of sufficient experiments and discussions are conducted in Section 3. Finally, we conclude this paper in the last section.
\section{Proposed Method}
\label{sec:method}
\begin{figure*}[htb]
\centering
\includegraphics[height=80mm,width=130mm]{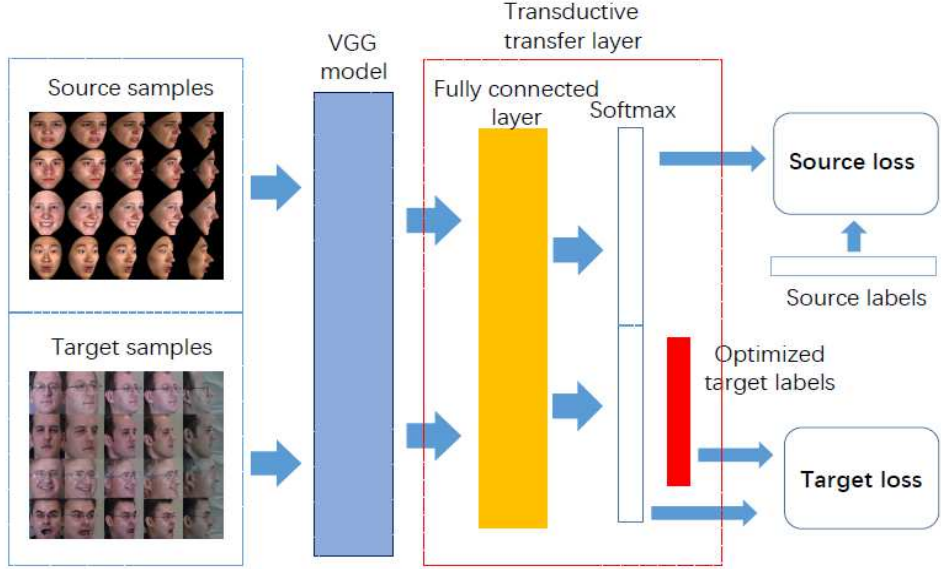}
\caption{The Training  Flow Chart of The Proposed Deep Transductive Transfer Learning Networks}
\label{fig:fig1}
\end{figure*}
In this section, we introduce our deep transductive transfer learning framework based on VGGface16-Net for non-frontal FER problem in details. Figure~\ref{fig:fig1} shows the network structure of the proposed transductive deep transfer learning, which consists of two sections: one is non-frontal facial expression feature learning part based on VGGface16-Net, the other is transductive transfer learning layer.
\subsection{Notations}
In order to facilitate the discussion of this paper, we first give some notations be used in the whole paper. We denote $\mathbf{X}^s=[x_1^s,x_2^s,...,x_{N_s}^s]$ as a set of as source domain samples from non-frontal facial expression database and $\mathbf{L}^s = [l^s_1,l^s_2,...,l^s_{N_s}]$ is the real class label vector set corresponding to $\mathbf{X}^s$, in which $x_i^s$ represents $i$th raw facial expression image sample and $N_s$ is the number of source database samples. Moreover, let $\mathbf{X}^{t}=[x_1^t,x_2^t,...,x_{N_t}^t]$ be the target instances set from the non-frontal facial expression database, in which $x_i^t$ represents $i$th image sample of target database and $N_t$ is the number of target database samples. We predefined $\mathbf{P}^t = [p^t_1,p^t_2,...,p^t_{N_t}]$ as network parameter which is the unknown class label vector set corresponding to $\mathbf{X}^{t}$ and is optimized with the update of network. In this model, each class label vector of $\mathbf{L}^s$ and $\mathbf{P}^t$ is represented as a $c\times1$ vector, in which $c$ is the number of facial expression classes and the all elements take the value of 1 or 0 in each class label vector of $\mathbf{L}^s$. Let $l^s_i=[l^s_{i,1},l^s_{i,2},...,l^s_{i,c}]$ be the label vector of $i$th sample, its each element satisfies the following rule to take value:
\[l^s_{i,j}=\begin{cases}
1,&\text{$x^s_i \in j$th class};\\
0,&\text{otherwise}.
\end{cases}j=1,...,c\]
Compared with the traditional subspace learning method, the ability of nonlinear representation is the advantage of deep neural networks. In general, the whole neural network can be regarded as a nonlinear mapping function, here we define this nonlinear mapping function as $\mathbf{f}(\cdot)$.
\subsection{Transductive Deep Transfer Learning Model}
Transductive transfer learning is a challenge topic because the target database samples have no labeled information to be utilized. In this paper, we follow the idea of the transductive transfer subspace learning model to focus on the cross-database non-frontal FER problem. Different from subspace learning theory in~\cite{zheng2016cross}~\cite{fernando2013unsupervised} et al., for learning more discriminative expression information between the source and target domain, we train a deep neural network model to eliminate the differences of feature distributions between the source and target samples as well as the discrepancy of distribution view-points intra-databases. For this purposes, we design a transductvie deep transfer learning framework (TDTL) which consists of one feature learning section and one transductive transfer learning layer. In the first section, we adopt a contemporary widely-used VGGface16-Net to deal with the raw non-frontal facial expression images. Remarkably, this choice is based on two reasons: one is that the VGGface16-Net can effectively extract very excellent facial expression features and acquire useful adaptive transfer knowledge, the other is that the performance of VGGface16-Net in classification task of transfer learning is better than other state-of-the-art deep neural networks such as AlexNet~\cite{krizhevsky2012imagenet} and GoogLeNet~\cite{szegedy2015going}. The VGGface16-Net is a deep neural network model which has five stacks of convolution network, plus three fully-connected layers, with a total of 16 layers. In deep neural network framework, this deep structure can help us to acquire highly sophisticated cross-database non-frontal facial expression features effectively. In the feature learning section of the TDTL model, we retain the five stacks of ConvNet and the first two fully-connected layers of the VGGface16-Net to extract the feature of facial expression. Five stacks of VGGface16-Net consist of thirteen convolutional layers in total, in which each stack is followed by one max pooling layer and each convolutional layer contains one activation function, that is rectified linear units (ReLU):
\begin{equation}\label{eq1}\nonumber
 f(x)=max(0,x).
\end{equation}
It is worth mentioning that these convolution layers in the VGGface16-Net use many smaller convolution kernels ($3\times3$ or $1\times1$ ) different from many other state-of-the-art deep neural network models which contain convolution layer with larger convolution kernel ($5\times5$ or $7\times7$ ). More small convolution kernels are equivalent to more nonlinear mapping, which can increase the representation ability of network and extract more discriminative non-frontal facial expression feature. In addition, smaller convolution kernels can significantly decrease the number of network parameters and improve operational efficiency of neural network. Follow the five stacks of ConvNet, there are two fully-connected layers which transform these nonlinear low-level description features of the raw non-frontal facial expression images into high level semantic information. In the second section of the TDTL model, we design a novel architecture of transductive transfer learning layer which includes one fully-connected layer and one softmax layer to further learn the higher semantic information for our classification task of cross-database non-frontal facial expression, in which the fully-connected layer uses the hyperbolic tangent function $tanh(\cdot)$ as its nonlinearity activation function. The softmax layer is used to accomplish the finally classification work.
\subsection{Transductive Deep Transfer Learning Training}
According to the general tranductive transfer learning method, in the TDTL model, the source domain samples are used as the training data set and the target domain samples are used as the testing data set. These two data sets are merged together and then divided into different batches. Every batch contains the source samples and target samples according to certain proportion. In the TDTL network, the training and testing are conducted step by step according to one batch and then another batch. To obtain better recognition results, we define a special loss function in accordance with  the proposed transfer learning networks framework as follows :
\begin{equation}\label{eq1}
  \mathcal{L}=\lambda_1\mathcal{L}_1+\lambda_2\mathcal{L}_2
\end{equation}
in which $\mathcal{L}_1$ and $\mathcal{L}_2$ are two different regularization terms to harmonize the facial expression feature learning and classification task, $\lambda_1$ and $\lambda_2$ are trade-off parameters to balance the two regularization terms $\mathcal{L}_1$ and $\mathcal{L}_2$. In $\mathcal{L}$, the first regularization term $\mathcal{L}_1$ is cross entropy loss function that depicts the distance of the training samples between the actual output (probability value) and the expected output (actual value).
\begin{equation}\label{eq1}
  \mathcal{L}_1=-\sum_{i=1}^{b_s}\sum_{j=1}^{c}l^s_{i,j}\mathbf{ln}y_{i,j}
\end{equation}
in which $b_s$ is the number of the training samples in every batch, the subscript $s$ denotes that the training samples are from the source database, $y_i=[y_{i,1},...,y_{i,c}]^T$ represents the prediction label value vector of the $i$th training sample and $l^s_i=[l^s_{i,1},l^s_{i,2},...,l^s_{i,c}]^T$ is real label value vector of $i$th the training sample. $y_{i,j}$ is calculated by a softmax function:
\begin{equation}\label{eq1}
  y_{i,j}=\frac{e^{o^s_{i,j}}}{\begin{matrix} \sum_{j=1}^c e^{o^s_{i,j}} \end{matrix}},  ~~~~ \forall j=1,...,c  \nonumber
\end{equation}
in which $o^s_i$ represents the output of network corresponding to $i$th training sample $x^s_i$, namely
$o^s_i = \mathbf{f}(x_i^s)$. In addition,
 the second regularization term $\mathcal{L}_2$ is the proposed transductive transfer learning loss function:
\begin{equation}\label{eq2}
  \mathcal{L}_2=\sum_{i=1}^{b_t}\Vert{p}_i^t-\mathbf{f}(x_i^t))\Vert_2^2 + \alpha\sum_{i=1}^{b_t}\Vert{p}_i^t\Vert_1
\end{equation}
where the $b_t$ is the number of the testing samples in one batch and the subscript $t$ denotes that the testing samples come from the target database. The second term of $\mathcal{L}_2$ is a $l_1$ norm regularization term which can ensure the sparse structure of the predicted label values matrix $\mathbf{P}^t$, $\alpha >0$ is trade-off parameter to control the sparsity of the columns of $\mathbf{P}^t$. When the value of $\alpha$ is larger, the each column of $\mathbf{P}^t$ will become more sparse than the value of $\alpha$ is smaller. More sparse means that the value of more elements is equal to 0, which can better accomplish classification tasks. Moreover, the network weights of each layer are updated according to the optimal value of the loss function by using the back propagation algorithm. The final task of the TDTL model is to predict the label value matrix of the $\mathbf{P}^t$ from the target domain samples which are used as testing samples.
\section{Experiments}
\label{sec:exp}
\subsection{The choice of samples}
\begin{figure*}[htb]
\centering
\includegraphics[height=40mm,width=160mm]{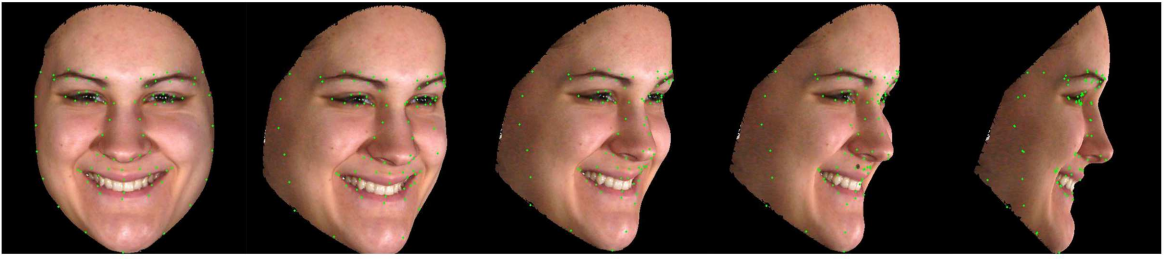}
\caption{The located 68 landmark points for SIFT features in 5 viewpoints}
\label{fig:fig2}
\end{figure*}
\label{sec:samples}
\begin{figure*}[htb]
\centering
\includegraphics[height=40mm,width=160mm]{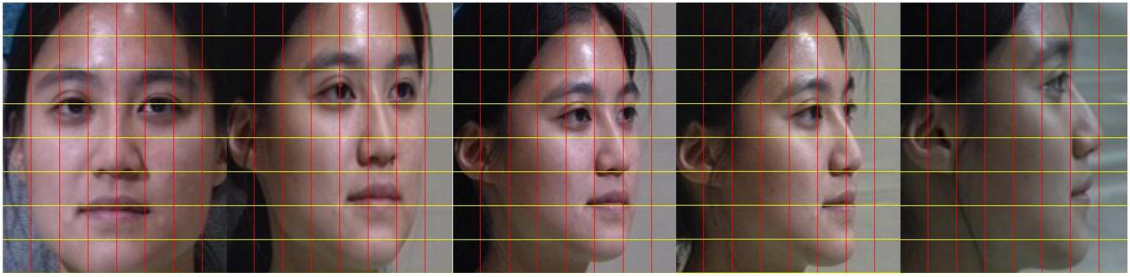}
\caption{The selected facial blocks ($8\times8$) for LBP feature extraction in 5 viewpoints}
\label{fig:fig3}
\end{figure*}
In this section, we conduct extensive experiments based on cross-database non-frontal facial expression images to evaluate the proposed transductive deep transfer learning model. We adopt two widely-used multi-view facial expression databases: the Carnegie Mellon University multi-pose, illumination, and expression (Multi-PIE) face database and the Binghamton University 3D Facial Expression (BU-3DEF) database in our experiments. The Multi-PIE database is a classic database developed by Gross et al.~\cite{gross2010multi} for non-frontal facial expression recognition, which was collected from 337 people. The images of Multi-PIE database include six basis facial expressions, such as disgust (DI), smile (SM), squint (SQ), scream (SC), surprise (SU), and neutral (NE) under 19 illumination environments and 15 viewpoints. The BU-3DEF database is established by Yin et al.~\cite{yin20063d} for 3D non-frontal facial expression classification, which is composed of 606 facial expression sequences collected from 100 subjects. This database contains seven fundamental expression categories, i.e., anger (AN), disgust (DI), fear (FE), neutral (NE), happiness (HA), sadness (SA), and surprise (SU) with multiple expression intensities. By comparing the expression categories of two databases, we select four common facial expressions (DI, SM/HA, SU, and NE) and 5 conventional viewpoints ($0^\circ$, $30^\circ$, $45^\circ$, $60^\circ$ and $90^\circ$) for our experiments from these two database, respectively. In addition, we randomly selected 100 subjects from all the 337 subjects of the Multi-PIE database, and for the BU-3DEF database, we choose all 100 subjects. It must be noted that, there are four expression intensities in three expression (HA, SU and DI) samples, while the NE expression only has one intensity in the BU-3DEF database. To the BU-3DEF database, we choose samples of HA, SU and DI expressions under five viewpoints with four expression intensities, and select samples of NE expression under five viewpoints with one expression intensities from 100 subjects, with a total of 6500 samples. Unlike BU-3DEF database, each type of expression of the Multi-PIE database only havs one expression intensity, and we choose samples of NE, SM, SU, and DI under five viewpoints with only expression intensity and one certain illumination condition from 100 subjects, in number 2000 samples. These samples will be used uniformly in the TDTL method and all comparison methods.
\begin{table}[!t]
\centering
\caption{The sample constitutions of the selected Multi-PIE and BU-3DEF databases with the same facial expression labels.}
\begin{tabular}{|c|c|c|}

\hline
Class& Multi-PIE & BU-3DEF\\
\hline
Disgust & 500 & 2000\\
\hline
Smile/Happiness & 500 & 2000\\
\hline
Surprise & 500 & 2000\\
\hline
Neutral & 500 & 500\\
\hline
Total & 2000 & 6500\\
\hline
\end{tabular}
\end{table}
\subsection{Experiments Based on Transductive Deep Transfer Learning}

\subsection{Experiments Based on Transductive Deep Transfer Learning}
The experimental protocol of the TDTL model is set according to the conventional transductive transfer learning method, namely our experiments are carried out when the source database is used as training samples set and the target database is used as testing samples set. When one of the BU-3DEF database or Multi-PIE database is used as the source database, respectively, the other is served as the target database. After extracting non-frontal facial expression image features, two full connected layer are exploited to learn network weights that can better represent transfer knowledge, and then we get a fixed 4096 dimension features from the two fully-connected layer as high-lever semantic information for classification task. The end of the network is the transductive transfer learning layer that includes one 4 dimension fully-connected layer and one 4-class softmax layer, which are used to recognize the facial expression categories of target database samples. In the TDTL model, the all input data are fixed-size 224 $\times$ 224 RGB raw facial expression images. In order to keep the optimization algorithm robustness and optimization efficiency, in the first two fully-connected layers based on VGGface16-Net model, the dropout ratio is 0.5, and the learning rate is set at 0.01. At the same time, the initialization network weights are sampled in Gauss distribution $\mathcal{N} (0, 0.01)$, and the bias item is initialized to 0. We use a min-batch size of 500, in which contains fifty percent proportion of the source and target samples respectively. Moreover, in the fully-connected layer of the transductive transfer layer, we start with a learning rate of 0.005, the dropout ratio is set at 50$\%$
. To the loss function of the TDTL model: $\mathcal{L}=\lambda_1\mathcal{L}_1+\lambda_2\mathcal{L}_2$, we alternately set the trade-off parameters $\lambda_1$ and $\lambda_2$ to 0 or 1 to optimize the parameters of network model. To $\mathcal{L}_2$, the trade-off parameter $\alpha$ is set to 150, in particular, we first randomly initialize the label matrix of target samples $\mathbf{P}^t$. The $\mathbf{P}^t$ is updated in pace with the parameters of the neural network until convergence of the loss function $\mathcal{L}$. The recognition accuracies are calculated through the predicted label values of $\mathbf{P}^t$ and the corresponding actual label values of $\mathbf{X}^t$.
\subsection{Comparison Experiments Setting}
For the purpose of the comparison, we choose recently proposed well-performing cross-database FER methods in dealing with the cross-database non-frontal facial expression classification problems including TTRLSR, SA(Subspace Alignment)~\cite{fernando2013unsupervised}, GFK(Geodesic Flow Kernel)~\cite{gong2012geodesic}, TKL(transfer kernel learning)~\cite{long2015domain}, TCA(Transfer Component Analysis)~\cite{Pan2011Domain}. In the SA approach, the source and target domains are jointly represented by seeking a optimal domain adaptation solution for learning a mapping subspace which aligns the source samples and the target one. The GFK method is transfer learning method based on manifold transform and kernel learning. The TKL algorithm can bridge the discrepancy of source and target distributions in the reproducing kernel Hilbert space based on a domain-invariant kernel schema. The key innovation of TCA is to minimize the distribution discrepancy in different domains based on the maximum mean difference theory. It is worth mentioning that, in FER problem, many previous FER works show that the recognition results based on hand-craft features are better than the raw image samples as input data in many traditional pattern classification methods~\cite{zheng2016cross},~\cite{Zong2018Learning},~\cite{zhao2007dynamic},~\cite{shan2009facial},~\cite{deniz2011face}.
For more reasonable comparison these five baseline methods, we select two classical hand-craft features (SIFT and LBP) for these comparison experiments although we directly use the raw image samples in experiments of the TDTL method. To measure the impact of features on the recognition results, we furthermore adopt VGG features of the samples from the BU-3DEF and Multi-PIE databases for our comparison experiments.

To extract SIFT features of the BU-3DEF and Multi-PIE databases, we first use OpenGL software to capture 2D non-frontal facial expression image samples from 3D facial expression models of the BU-3DEF database. Before extracting the SIFT features, we manually locate the 68 landmark points for each facial image, in which these landmark points (see Fig.~\ref{fig:fig2}) as the key points for SIFT feature extraction are located in the major parts of AUs including mouths, brows, eyes, noses and face contour. According to the extraction method of SIFT feature, the SIFT feature of each sample is in size $68\times128$. We furthermore transformed the each extracted SIFT feature into a vector of length 8704($68\times128$). Different from SIFT feature, we apply a LBP operator the 59-bin $LBP_{8,1}^{u2}$ to extract LBP descriptors of the BU-3DEF and Multi-PIE databases, in which the subscript $(8,1)$ indicates adopting the operator in a $(8,1)$ neighbourhood and the superscript $u2$ represents using only uniform patterns and labelling all remaining patterns with a single label. Each facial image was divided into 64$(8\times8)$ (see Fig.~\ref{fig:fig3}) regions and represented by the LBP histogram of these regions with the vector length of 3776$(64\times59)$. Moreover, we adopt the same VGGface16-Net model as our method to extract very efficient deep neural network feature for our comparison experiments. The extracted VGG feature of each sample is a vector of length 4096.
\begin{table*}[!t]
\centering
\caption{Experimental results of all methods on BU-3DEF and Multi-PIE Databases According To Recognition Accuracy And F1-score .}
\scalebox{1}[1]{
\renewcommand{\arraystretch}{1.3}
\begin{tabular}{|l|c|c|c|c|}
\hline
\multirow{2}{*}{Method}&\multicolumn{2}{c|}{BU-3DEF to Multi-PIE}
&\multicolumn{2}{c|}{Multi-PIE to BU-3DEF} \\ \cline{2-3}\cline{4-5}
{}                 & Accuracy (\%) &F1-score &  Accuracy (\%) &F1-score\\ \hline
SIFT (68) + SA     & 33.50         &0.2571   & 33.38          &0.3102  \\ \hline
SIFT (68) + GFK    & 31.82         &0.2177   & 35.05          &0.3109  \\ \hline
SIFT (68) + TKL    & 36.95         &0.3100   & 34.80          &0.2855  \\ \hline
SIFT (68) + TCA    & 39.20         &0.3767   & 38.63          &0.3209  \\ \hline
SIFT (68) + TTRLSR & 41.30         &0.3829   & 40.66          &0.3521  \\ \hline  \hline
LBP (8*8) + SA     & 45.60         &0.3707   & 33.18          &0.2127  \\ \hline
LBP (8*8) + GFK    & 44.20         &0.3859   & 34.92          &0.2536  \\ \hline
LBP (8*8) + TKL    & 33.85         &0.4208   & 33.00          &0.2743  \\ \hline
LBP (8*8) + TCA    & 44.20         &0.3550   & 34.92          &0.3274  \\ \hline
LBP (8*8) + TTRLSR & 43.90         &0.4137   & 31.07          &0.3219  \\ \hline  \hline
VGG + SA           & 55.00         &0.5275   &50.82           & 0.4754 \\ \hline
VGG + GFK          & 53.75         &0.4976   &57.80           & 0.5165 \\ \hline
VGG + TKL          & 53.90         &0.4842   &50.46           & 0.4774 \\ \hline
VGG + TCA          & 57.80         &0.5545   &58.32           & 0.5228 \\ \hline \hline
fine-tuned VGG     & 55.30         & -        &50.20           &-\\ \hline
TDTL               &\textbf{66.85} &\textbf{0.6547}   &\textbf{66.05}  &\textbf{0.5309}  \\ \hline
\end{tabular}
}
\label{tab:tab1}
\end{table*}
The detailed parameters setting of all baseline methods in these experiments are reveal as follows:
\begin{enumerate}
\item
For the SA method, we traverse the subspaces of dimensionality for $d=[1:300]$ with interval 1 and select the optimal recognition rates and record them in the Table~\ref{tab:tab1}.
\item
For the TKL method, we large range search the parameter $\eta$ from [0.01:0.01:0.09], [0.1:0.1:1] and [2:1:15], and the linear kernel is adopted as kernel function in this method, the best recognition results of three features are recorded in the Table~\ref{tab:tab1}.
\item
For the GFK method, we use one kernel-based method: GFK(PCA, PCA) and traversing the dimensionality of PCA from 10 to 100 with interval 10, we search the best results to recorded in the Table~\ref{tab:tab1}.
\item
For the TCA method, we large scope search the trade-off parameter $\mu$ from a preset parameter interval [1:300] and utilize the monomial linear kernel to calculate the kernel matrix of the TCA algorithm. The best results of corresponding to the various features are recorded in the Table~\ref{tab:tab1}.
\item
For the TTRLSR method, we use grid search strategy to search the parameter $\lambda$, $\alpha$ and $\tau$, and the parameter
grid is arranged at [0.01:0.01:0.09], [0.1:0.1:0.9] and $[1:100]$ with interval 5. Furthermore, we have tried to traverse
the size of auxiliary data set from whole target data set according to 10\% to 100\% with interval 10\%. Finally, we choose the best results from different proportion and parameters.
\end{enumerate}
\subsection{Experimental Results and Discussions}
\begin{figure}[htb]
\centering
\includegraphics[height=60mm,width=70mm]{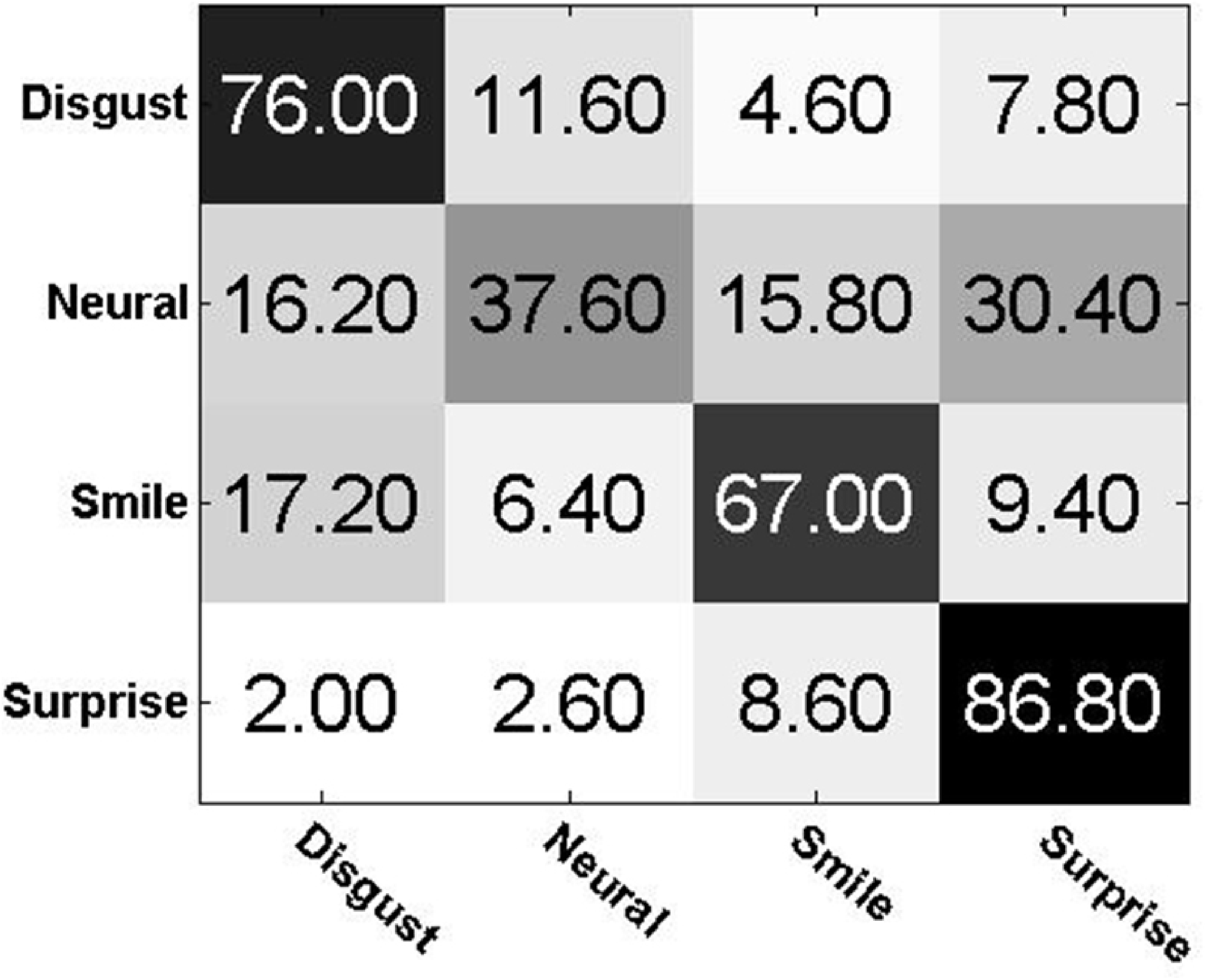}
\caption{The recognition rate confusion matrix of the TDTL method use BU-3DEF database as source samples and Multi-PIE database as target samples.}
\label{fig:fig4}
\end{figure}
\begin{figure}[htb]
\centering
\includegraphics[height=60mm,width=70mm]{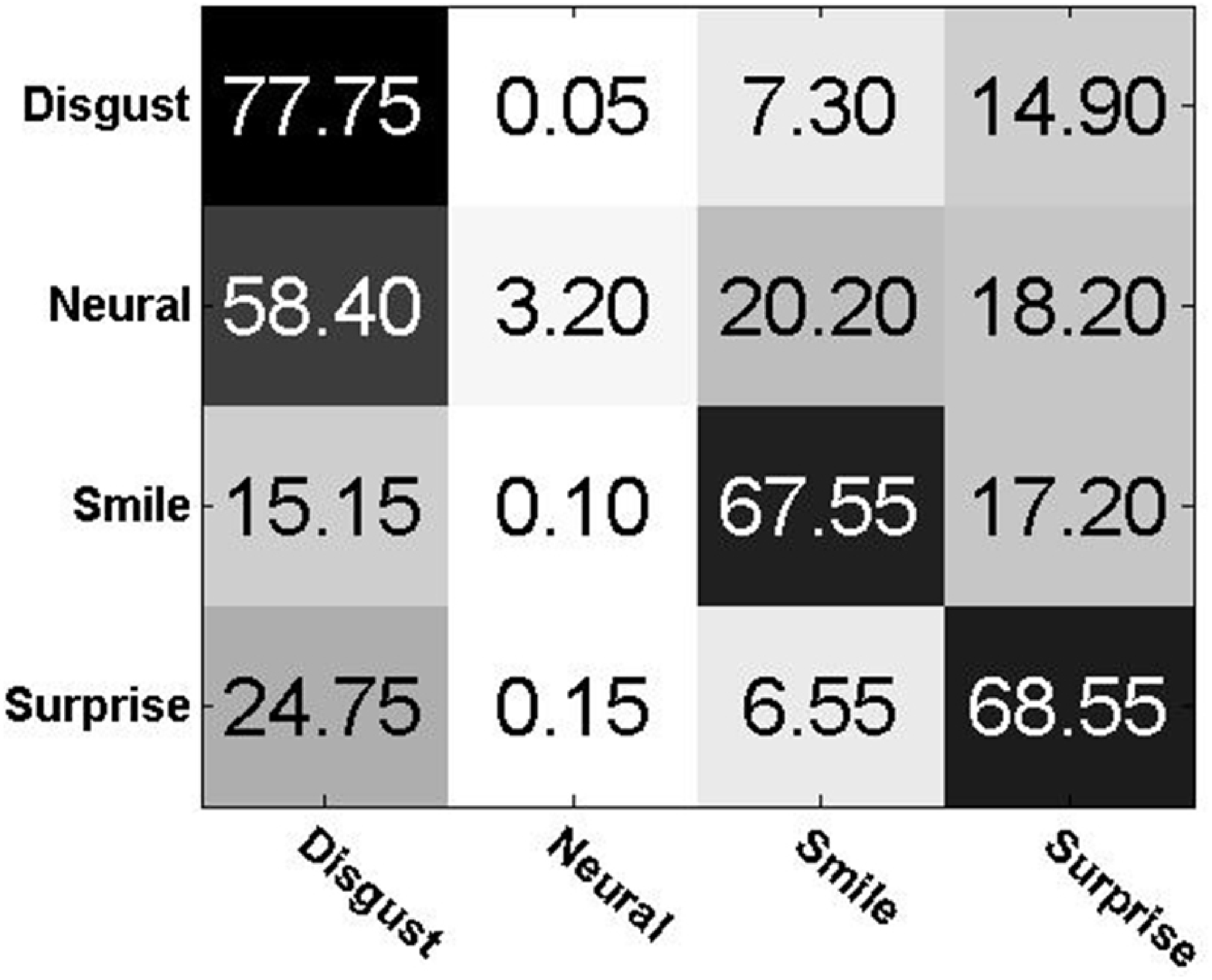}
\caption{The recognition rate confusion matrix of the TDTL method use Multi-PIE database as source samples and BU-3DEF database as target samples.}
\label{fig:fig5}
\end{figure}
In this section, we will report the all recognition accuracies of the TDTL method and all comparison results based on recent the state-of-the-art transfer learning methods. These experimental results according to recognition accuracy $(\%)$ are showed in Table~\ref{tab:tab1}. The recognition accuracy is calculated in term of $\frac{P_r}{T_e}\times 100$, where $P_r$ is the number of correct predictions to target domain samples and  $T_e$ is the total number of target domain samples. In addition, according to Section~\ref{sec:samples} the BU-3DEF database is imbalanced according to the category in our experiments. To more objectively reflect the performance of all methods in this paper, we furthermore report the F1-score $(\frac{1}{c}\sum_{k=1}^d\frac{2p_k\times r_k}{p_k + r_k})$ of all the experimental results, in which $p_k$ and $r_k$ express the precision and recall of the $k$th facial expression, respectively, and $c$ means the number of facial expression categories. From table~\ref{tab:tab1}, we can see that the TDTL model has achieved better recognition accuracies $(66.05\%$ $\&$ $66.85\%)$ than these comparison experimental methods whether the BU-3DEF database is used as source domain samples or the Multi-PIE database as source domain samples. It's also worth mentioning that, the F1-score of our method have also better performance (0.6547 $\&$ 0.5309) than other comparison experimental methods. To summarize, from the performances of recognition rates and F1-score, the proposed TDTL method is more suitable for dealing with cross-database non-frontal FER problem between the BU-3DEF and Multi-PIE databases. Moreover, from the comparison experimental results based on three features, it is also distinctly to see that the recognition accuracies of the VGG features are better than the SIFT and LBP features, meanwhile, the comparison experimental results of SIFT and LBP features are not significantly different in all comparison methods. In general, these comparison experimental results indicate that the VGG features can better represent the complicated information such as non-frontal facial expression data than the traditional hand-craft features like SIFT and LBP features. It is worth mentioning that, the TTRLSR model~\cite{zheng2016cross} also has achieved good recognition results and F1-score, especially in the SIFT features compared with other comparison methods. This phenomenon shows that transductive transfer learning method based on group sparse learning method can also achieve good results in complex cross-database non-frontal facial expression recognition tasks. In addition, the TTRLSR method need to select feature channels of the samples, the VGG feature does
not satisfy this characteristic, thus the TTRLSR method has only the experimental results of the SIFT and LBP features. Finally, we also display the recognition rate confusion matrices of the TDTL method in Figs.~\ref{fig:fig4} and~\ref{fig:fig5}. Compared with Figure~\ref{fig:fig4} and Figure~\ref{fig:fig5}, we can clearly find that the DI, SM and SU expressions are more easily recognized, by contrast, the NE expression is more difficulty recognized by the TDTL method, especially when the BU-3DEF is used as a target database. This may be that the distribution of NE expression is similar to the distributions other three expressions, which leads to the low recognition performance of NE expression.
.

\section{Conclusions}
\label{sec:conclusion}
In this paper, a novel tranductive deep transfer learning (TDTL) framework based on widely-used VGGface16-Net is proposed to better deal with cross-database non-frontal FER problem. In this method, we designed a special transfer learning layer to jointly optimize the loss function $\mathcal{L}$ for predicting the label values of the target samples. To evaluate the TDTL method, extensive experiments are conducted on two publicly available non-frontal facial expression databases, i.e., BU-3DEF and Multi-PIE database. The experimental results demonstrate that the TDTL model can effectively enhance the recognition effects in coping with non-frontal FER problem compare with recent state-of-the-art transfer learning methods. Additionally, From the results of comparison experiment, we can see that the VGG-based features achieves more excellent recognition accuracies than the traditional hand-craft features on two databases. These results furthermore indicate that the deep neural network is more prominent in acquiring the feature representation of human facial emotion.

\end{document}